\RequirePackage{fix-cm}
\documentclass[smallextended]{svjour3}

\smartqed
\usepackage{dirtytalk}
\usepackage{ragged2e}
\usepackage{graphicx,xcolor}
\usepackage{amssymb,amsmath}
\usepackage{url}
\usepackage[square,sort,comma,numbers]{natbib}
\usepackage[english]{babel}
\usepackage{authblk}
\usepackage{anysize}
\usepackage{multirow}
\usepackage{subcaption}
\usepackage{soul}

\journalname{International Journal of Computer Vision}

\begin{document}

\title{VICSOM: {VI}sual {C}lues from {SO}cial {M}edia for psychological assessment}


\author{Mohammad Mahdi Dehshibi         \and
        Gerard Pons        \and
        Bita Baiani         \and
        David Masip
}

\authorrunning{Dehshibi et al.}

\institute{Mohammad Mahdi Dehshibi, Gerard Pons, and David Masip \at
              Department of Computer Science, Universitat Oberta de Catalunya, Barcelona, Spain \\
              \email{\{mdehshibi,gponro,dmasipr\}@uoc.edu} \\
              Bita Baiani \at
              Department of Psychology, Islamic Azad University, Science and Research Branch, Tehran, Iran \\
            }

\date{Received: date / Accepted: date}

\maketitle

\begin{abstract}
Sharing multimodal information (typically images, videos or text) in Social Network Sites (SNS) occupies a relevant part of our time. The particular way how users expose themselves in SNS can provide useful information to infer human behaviors. This paper proposes to use multimodal data gathered from Instagram accounts to predict the perceived prototypical needs described in Glasser's choice theory. The contribution is two-fold: (i) we provide a large multimodal database from Instagram public profiles (more than 30,000 images and text captions) annotated by expert Psychologists on each perceived behavior according to Glasser's theory, and (ii) we propose to automate the recognition of the (unconsciously) perceived needs by the users. Particularly, we propose a baseline using three different feature sets: visual descriptors based on pixel images (SURF and Visual Bag of Words), a high-level descriptor based on the automated scene description using Convolutional Neural Networks, and a text-based descriptor (Word2vec) obtained from processing the captions provided by the users. Finally, we propose a multimodal fusion of these descriptors obtaining promising results in the multi-label classification problem.

\end{abstract}

\section{Introduction}
The complexity of the human mind can manifest through a dynamic and organized set of characteristics which uniquely influences the environment, cognition, emotions, motivations, and behaviors in various situations. These characteristics, which can disclose how people are individually different, are known as personality~\cite{friedman1999personality}. Not only psychologists but also sociologists and humanities researchers are also interested in knowing more about human personality and have been in collaboration with computer and data scientists to find computational models of personality trait inferences at different levels (\cite{rojas2010automatic,ponce2016chalearn}). On the other hand, analyzing the complex and subconscious behavior of humans has an impact on health, security, human-computer/machine/robot interaction, and even entertainment. The emergence of social network sites (SNS) provides a massive amount of visual and multimodal information and helps researchers to recognize clues associated with the subconscious behavior and situations of their users.

Social network sites enable individuals to construct and display their identities in favor of interacting with other members~\cite{boyd2007social}. Therefore, many individuals have increasingly invested in developing an idealized online self that they can present to the world~\cite{gonzales2011mirror}. Instagram is an image-based SNS and a simple way to capture and share life's moments, and follow friends and family to see their interests. Instagram allows users to upload photos and videos to the service, write captions, add tags, and location information. The service also supports messaging features, the ability to include multiple images or videos in a single post, as well as ``Stories", i.e. temporary posts that disappear after 24 hours. 

This pool of data, as a mirror of society on a smaller scale, can provide valuable clues about users' physical/mental health conditions, as well as personality features which have recently been used in analytical screening. Analytical screening methods have successfully identified markers in social media data, and this trend has been followed up in two directions including (i) physical ailments, and (ii) mental health issues, e.g., addiction~\cite{moreno2012associations}, depression~\cite{de2013predicting, katikalapudi2012associating, reece2017instagram}, Post-Traumatic Stress Disorder (PTSD)~\cite{harman2014measuring}, suicidal ideation~\cite{de2016discovering}, sense of love~\cite{persson2017love}, happiness~\cite{dodds2011temporal}, and enhance life satisfaction~\cite{oh2014does}. However, studies in health screening using SNS data are not mature enough and need to be developed in order to be effectively used in health care systems. Supporting pieces of evidence for this claim are (i) lack of publicly available data sets, (ii) lack of ground-truth, provided by psychologists, and (iii) focusing on a specific physical or mental health problem which decreases the generalizability.

This research aims at studying how humans intrinsically contribute towards behavioral motivation (i.e., human needs) by sharing their interest on Instagram. The analysis of human needs has always been important for many psychologists who have put forward many efforts to characterize these needs and have proposed a structure for the treatment of their subjects according to this classification. Two well-defined psychological theories known as the Maslow's hierarchy of needs~\cite{maslow1943theory} and Glasser's choice theory~\cite{glasser1999choice} make the foundation of our study. Based on the Maslow's hierarchy of needs, human motivations generally move through ``physiological," ``safety," ``belonging and love," ``esteem," ``cognition," ``aesthetic," ``self-actualization," and ``transcendence" patterns, respectively. Indeed, the individual must be satisfied by each level to find enough motivation for thinking about at the next level and completing their hierarchy. 

\begin{figure}[htbp]
    \centering
    \begin{subfigure}[c]{0.45\textwidth}
        \includegraphics[width=\textwidth]{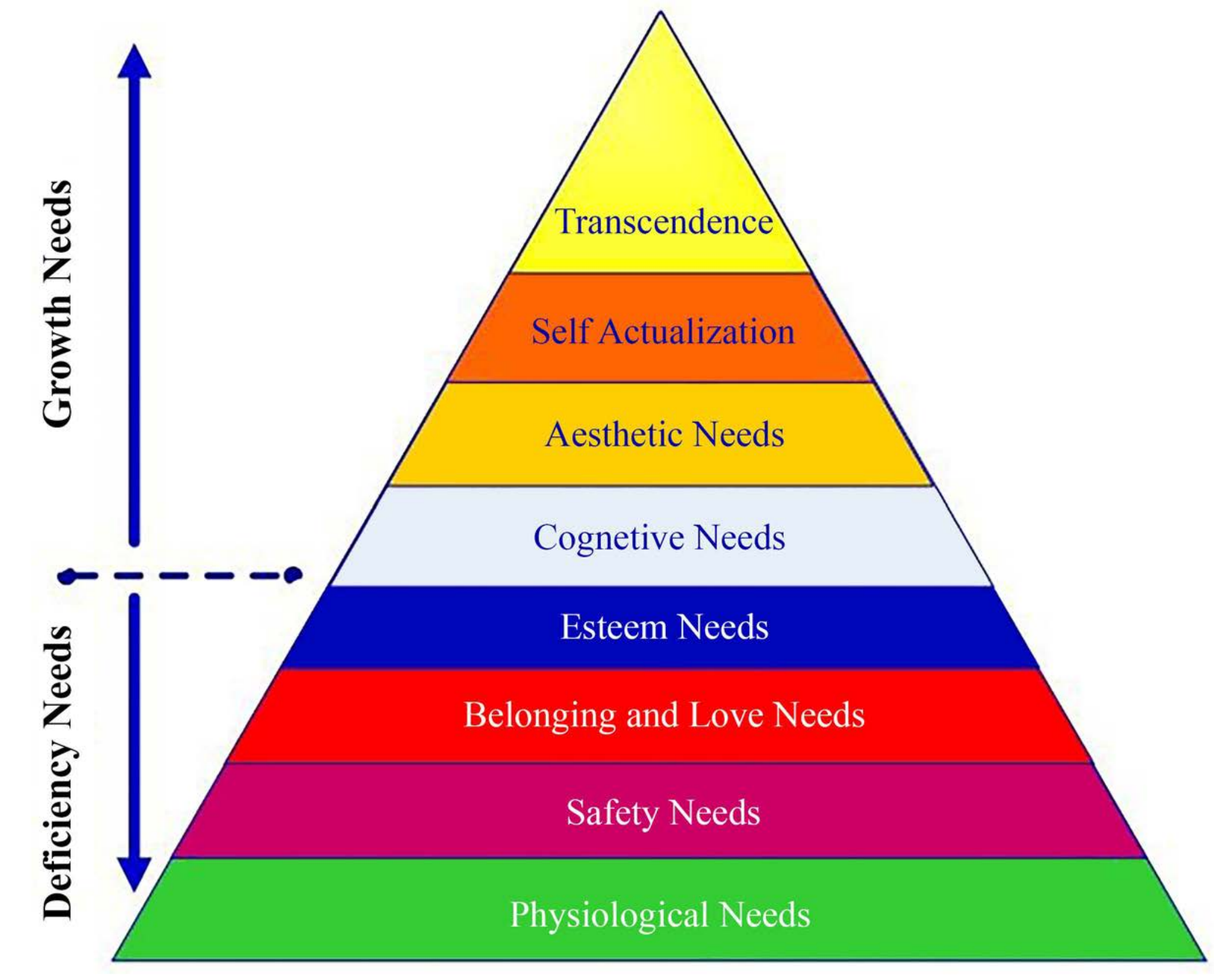}
        \caption{}
        \label{fig:M2a}
    \end{subfigure}
    \hfill
    \begin{subfigure}[c]{0.5\textwidth}
        \includegraphics[width=\textwidth]{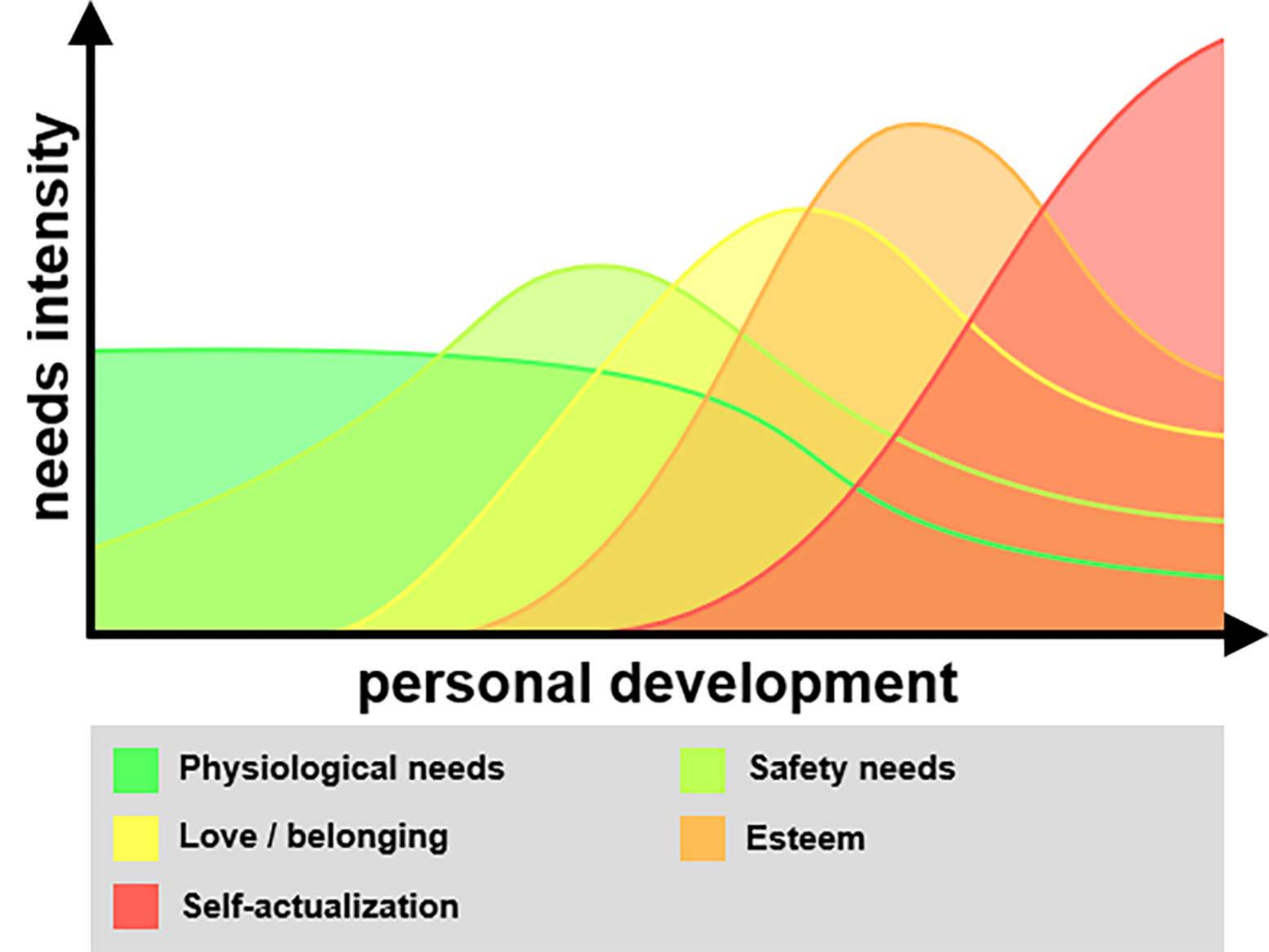}
        \caption{}
        \label{fig:M2b}
    \end{subfigure}
    \caption{Maslow's hierarchy of needs represented as~(a) pyramid ~\cite{maslow_motivation_1954}, and~(b) a dynamic hierarchy with overlaps of different needs at the same time~\cite{steere1988becoming}.}
    \label{fig:M2}
\end{figure}

This psychological theory is an infrastructure for understanding the correlation between drive and motivation in human behavior~\cite{deckers2018motivation,maslow_motivation_1954} because Maslow stated that \say{Instead of stating that the individual focuses on a certain need at any given time, it must be stated that a certain need `dominates' the human organism.} This well-defined hierarchy which is used in sociology research, management training, and secondary and higher psychology instruction, states that these levels overlap with each other. Fig.~\ref{fig:M2} shows the original hierarchy and its alternative illustration as a dynamic and overlapping hierarchy of needs.

Although the human complex brain can think about different phenomena in parallel, this theory states that if a human is struggling to meet their physiological needs, he might not be able to pursue safety, belongingness, esteem, and self-actualization. These concepts, associated with each level of needs, are abstract and need to be clarified by in-detail words and synonyms. For instance, physiological needs include homeostasis, food, water, sleep, shelter, and sex~\cite{maslow1943theory,maslow_motivation_1954,mak2001confirmatory}. Safety and Security needs include personal security, emotional security, financial security, and health and well-being~\cite{harries2008feeling,howell2013money,henwood2015maslow}. Social Belonging needs include friendships, intimacy, family~\cite{taormina2013maslow}. Self-actualization can include parenting, utilizing abilities, utilizing talents, pursuing a goal, seeking happiness~\cite{maslow2013toward}.

When it comes to analyzing posts from SNS, for some cases providing descriptions about higher levels in the pyramid is not possible. Consequently, to find some visual/textual clues for a better representation of the perception of human needs within shared media on Instagram, we restrict this research on Glasser's choice theory \cite{glasser1999choice}. Based on this, human behavior is driven by five categories of needs, namely: `Survival' (e.g. food, clothing, shelter,  personal safety, or sex),  `Belonging' (e.g. connecting, love), `Power' (e.g. significance, competence), `Freedom' (i.e. autonomy), `Fun' (i.e. learning).

In this research, we introduce a new database from Instagram user profiles enriched with ground-truth provided by an expert psychologist (following Glasser's choice theory). We propose an automated method for predicting the perception of human needs from Instagram profiles based on three informational cues: (i) visual information extracted from state-of-the-art computer vision image descriptors, (ii) high level descriptors from the scene contents (both in terms of scene categorization and object recognition), extracted using two Convolutional Neural Networks, and (iii) the processed textual information accompanying each image (captions). We also propose a multimodal fusion of the three signals, obtaining promising results in which is, to the best of our knowledge, the first contribution towards the automated analysis of perceived human needs using SNS data. Related work is surveyed in Section~\ref{survey}. Section~\ref{vicsom} dedicates to the description of the database, its organization, associated meta-data, and distribution conditions. Automatic recognizing the perceived subject's needs along with the analytical evaluations of the proposed methodology form the content of Sections~\ref{method} and~\ref{experiment}, respectively. Finally, concluding remarks are drawn in Section~\ref{conclusion}.

\section{Related work}\label{survey}
Sharing multimodal data (pictures, videos, texts) has become an essential part of the online social experience. This data can provide valuable clues about the physical, mental health conditions, personality features, characters, and needs of its users even if users are not yet aware that their health has changed. From another perspective, analyzing human needs can reveal individual motives for their behaviors. 

Different researchers have used SNS data to plan the path of disease occurrences~\cite{christakis2010social,reece2017forecasting}. Predictive screening methods have also successfully found signs of mental health issues in social media data~\cite{de2013predicting,weiser2015me,reece2017instagram}.

Reece et al.~\cite{reece2017instagram} used a computational model to predict depression signs in users' Twitter data and showed that screening the posts on Twitter can effectively identify this condition earlier and more accurately than the health professionals. Results of this study showed that depression indicators are identifiable within six months before the trauma appears in an individual. This progress, compared to the average 19-month delay between trauma event and diagnosis experienced by the individuals, can provide a framework for an accessible, accurate, and inexpensive depression screening, where in-person assessments are difficult or costly.

Kim and Kim~\cite{kim2018using} utilized computer vision approaches to find whether there is a positive relation between shared images characteristics and personality traits. The data consists of 25,394 photos shared over 179 Instagram profiles where the owners were university students. They measured user's characteristics with an online survey. Content categorization was done by counting the number of faces, analyzing the emotions on the faces, and the pixel derived features using Microsoft Azure Computer Vision API~\cite{microsoft}. Finally, they concluded that Instagram users' extraversion, openness, agreeableness, and conscientiousness are associated with the features of photos they have shared. Although this study put a step forward by analyzing the content, they stated that the photo, itself, is enough for concluding and they did not consider texts which usually appear in profile biography, captions, and comments. They believed that expressing oneself by photo is simple because the individual does not need to care much about word selection and grammatical errors. However, observing the contradiction between the image content and the written caption for it helps to discover some hidden parts of a person's mental state. Indeed, photos context, caption of posts, textual reply to comments, profile image, and profile biography are all clues that can provide an insight view to the mental state of a user.

Kircaburun and Griffiths~\cite{kircaburun2018instagram} examined the relationships between personality, self-liking, daily Internet use, and Instagram addiction. They asked 752 university students to complete a self-report survey, including the Instagram Addiction Scale and the Self-Liking Scale. They reported that agreeableness, conscientiousness, and self-liking are negatively associated with Instagram addiction while daily Internet usage is positively associated with Instagram addiction. However, the majority of shared contents on Instagram is not only about selfie and self-liking and users also tend to share personal interests through image, videos, and text over a photo, as well.

Pampouchidou et al.~\cite{pampouchidou2017automatic} surveyed methods published from 2005 to 2017 about automatic depression assessment based on visual cues. They addressed several research questions, including the number of modalities employed, facial signs, experimental protocols for dataset acquisition, feature descriptors, decision methods, and scores. They concluded that results are consistent with the social withdrawal, emotion-context insensitivity, reduced reactivity hypotheses of depression, and the importance of dynamic features/multimodal approaches through the quantitative analysis. They also mentioned that the multitude of reported approaches on automatic depression assessment is not mature enough because clinical research questions such as the capacity to distinguish between different depression sub-types or the influence of ethnicity and culture on the progress of mental health were not addressed systematically. Finally, they argued that visual cues need to be supplemented by information from other modalities to achieve clinically useful results.

Most of the studies covering social media analysis have targeted some specific personality disorder/traits. The foundation was created based on answers to an online questionnaire to reveal if the SNS user has a particular personality disorder. Therefore, apart from the truth level of the answers, unavailability of this information causes the contextual photo analysis to become meaningless. In this study, we resolve the mentioned shortcomings by introducing the VICSOM database which contains multimodal data of 86 Instagram profiles from both Persian and Spanish users with 30,080 photos. Moreover, we investigated the relationship between activity in Instagram and the perceived needs the individual seeks according to the Glasser's choice theory.

\section{VICSOM Database} \label{vicsom}
Instagram is a data pool in which we can perceive the user's needs, feelings, and thoughts by analyzing shared contents. Moreover, Instagram users can enrich this expression by adding textual and hashtag-based captions to images. In collecting the VICSOM, we targeted public accounts at the time of scraping from two regions, i.e., Iran from the Middle East and Spain from Europe. Recruitment and data collection procedures were identical for both regions. We made a one-time collection of participants' Instagram profile. In total, we collected 30,080 images from 86 Instagram users for the analysis of human needs. Commercial and celebrity profiles were ignored in the profile selection. We used the Instagram developer's Application Programming Interface (API) to harvest data from public pages.
 
The expert psychologist has then visited the pages and provided a description of the need-level of users based on Glasser's theory. Both Persian and English versions of these descriptions are available in the database. We also provided the labels for each profile according to the five categories in the Glasser's choice theory, i.e. each profile was labeled with $l \in \mathcal{P}(L)-\emptyset$, where $\mathcal{P}(L)$ is the power set of $L$ and $L=\{\text{survival}, \text{belonging}, \text{power}, \text{freedom}, \text{fun}\}$.
 
The expert psychologist also provided a description to label 32 profiles according to Maslow's hierarchy of need. This data is not used in the experiments, but it is available in the database.

\subsection{Data Statistics and subject demographics}
We collected data from 86 Instagram users, totaling 30,080 images. The mean number of posts per user was 286.47 (SD = 198.18). This distribution was skewed by a smaller number of frequent posters, as evidenced by a median value of just 286.47 posts per user. The subjects were coming from 2 different cultural backgrounds, i.e., Iranian and Spanish. 54 of the subjects are males, 32 are females, resulting in a gender ratio (male/female) of 1.68. See Table~\ref{tbl:M2} for summary of statistics.

\begin{table}[htbp]
\centering
\caption{Summary statistics and Demographics for VICSOM}
\label{tbl:M2}
\begin{tabular}{l|lll}
\hline
                       & Female & Male & Age range  \\ \hline
\multirow{1}{*}{Iran}      & 12     & 30   & 15-50            \\
                          
\multirow{1}{*}{Spain}    & 20     & 24   & 15-50         \\ \hline
\end{tabular}
\end{table}

VICSOM~\footnote{The database is available for research purposes upon request and EULA signature.} has two sets as follows:

\begin{itemize}
	\item \textbf{Set 1} contains 42 multimodal data of Iranian individuals who are the owner of public pages. Note that the profiles were public at the time when the data was acquired and, since the users can change their privacy settings, we can not ensure the profile is publicly available anymore.  Following the suggestion of expert psychologist and the use of previous experience~\cite{bastanfard2007iranian,dehshibi2010new,dehshibi2017cubic}, we decided to mine pages that look more realistic considering the current cultural, social, political, and economic conditions of Iran. The data is composed of a set of images (with a maximum of 1000) posted by the subject and a JSON file containing the caption of the photos including hashtags, and the geographical tags. 
    \item \textbf{Set 2} contains 44 multimodal data of Spanish individuals who are the owner of public pages. The mentioned situations for gathering Iranian Instagram profiles were also observed in selecting Spain profiles.
\end{itemize}

\begin{figure}[htbp]
    \centering
    \includegraphics[width=0.65\textwidth]{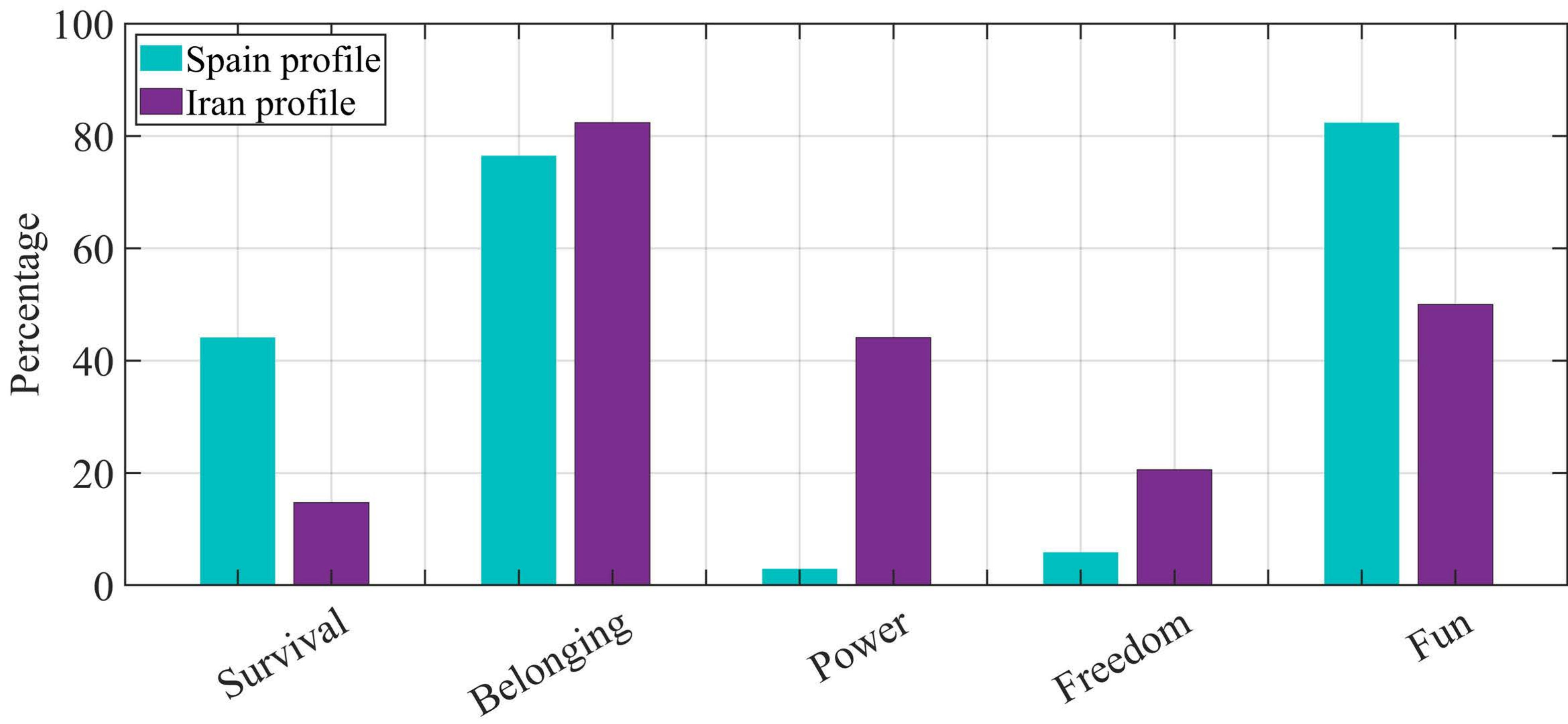}
    \caption{Diversity of labels per country.}
    \label{fig:M6}
\end{figure}

The expert psychologist visited each user's profile, and provided labels for them based on Glasser's choice theory. Therefore, one subject could be perceived as to be looking for any combination of all five needs. Fig.~\ref{fig:M6} shows the perceived needs distribution for the two countries (Iran and Spain). 

\section{Automatic recognition of subject needs using SNS multimodal data} \label{method}
In this section, we propose a baseline approach for classifying the visual/textual data obtained from the Instagram profiles of this database. In terms of machine learning, this is a multi-label classification problem given the fact that one or more categories can be perceived in the assessment of a subject. To learn a classification model, we used three different feature representation methods:

\begin{itemize}
    \item A feature space consisting of a bag of visual words using the SURF descriptor.
    \item A histogram of visual tags provided by two different Convolutional Neural Networks.
    \item Textual descriptors from captions using word2vec~\cite{word2vec} latent space.
\end{itemize}

We also explored a multimodal fusion of both visual and textual cues for the multi-label classification.

\subsection{Bag of Visual Words}
The idea of bag-of-visual-words (BoVW) \cite{csurka2004visual} was borrowed from natural language processing~\cite{zhang2010understanding} in which a histogram containing the frequency of word occurrence represents a document. In Bag of Visual Words, the image is the equivalent of the document, and the words are cluster centers of local descriptors. Different feature descriptors can be used, and salient point detectors (SIFT or SURF descriptors) have demonstrated their performance~\cite{juan2007comparison}. 

In this study, we used speeded-up robust features (SURF)~\cite{bay2006surf}. To compute the keypoints and descriptors, first, a square-shaped filter of size $8 \times 8$ is applied to the integral image to produce the Laplacian of Gaussians. Then, the Hessian matrix is calculated by using a blob detector to detect interest points. Given a point $p=(x, y)$ in an image $I$, the Hessian matrix $H(p, \sigma)$ at point $p$ and scale $\sigma$, is:

\begin{equation} \label{eq:hessian_matrix}
\mathbf{H(p, \sigma)} = \begin{bmatrix} D_{xx}(p, \sigma) & D_{xy}(p, \sigma) \\ D_{xy}(p, \sigma)v & D_{yy}(p, \sigma) \end{bmatrix}
\end{equation}
where $D_{\bullet}(p,\sigma)$ is the convolution of the second-order derivative of Gaussian with the image $I$ at the point $p$. The square-shaped filter of size $8 \times 8$ is an approximation of a Gaussian with $\sigma=1.2$ which represents the highest spatial resolution for blob-response maps. Afterwards a square window with the size of $20 \times 20$ is extracted, centered on the interest point and oriented along the orientation to describe the region around the point. Then, the interest region is split into smaller $4 \times 4$ square sub-regions, and the Haar wavelets with a size of $2\sigma$ are calculated for each one. This results in feature vectors containing 64 dimensions which are invariant to rotation, change of scale and contrast. In order to convert vector-represented patches into visual words and generate a representative dictionary, the vectors are clustered into $k$ (in this study $k =256$) groups using k-means~\cite{kanungo2002efficient} and the cluster centers consider as a vocabulary of $k$ visual words. SURF feature descriptor~\cite{bay2006surf} was applied to 30\% of all images to construct the visual vocabulary (BoVW). In our implementation, the block width is $[32~64~96~128]$, and 80 percent of the strongest features were kept, obtaining a feature vector $x \in \mathcal{X}_{BoVW} \subseteq \mathbb{R}^{256}$.

\subsection{CNN-based Bag of Words for context information extraction}
\begin{figure}[htbp]
    \begin{subfigure}[c]{0.4\textwidth}
        \includegraphics[width=\textwidth]{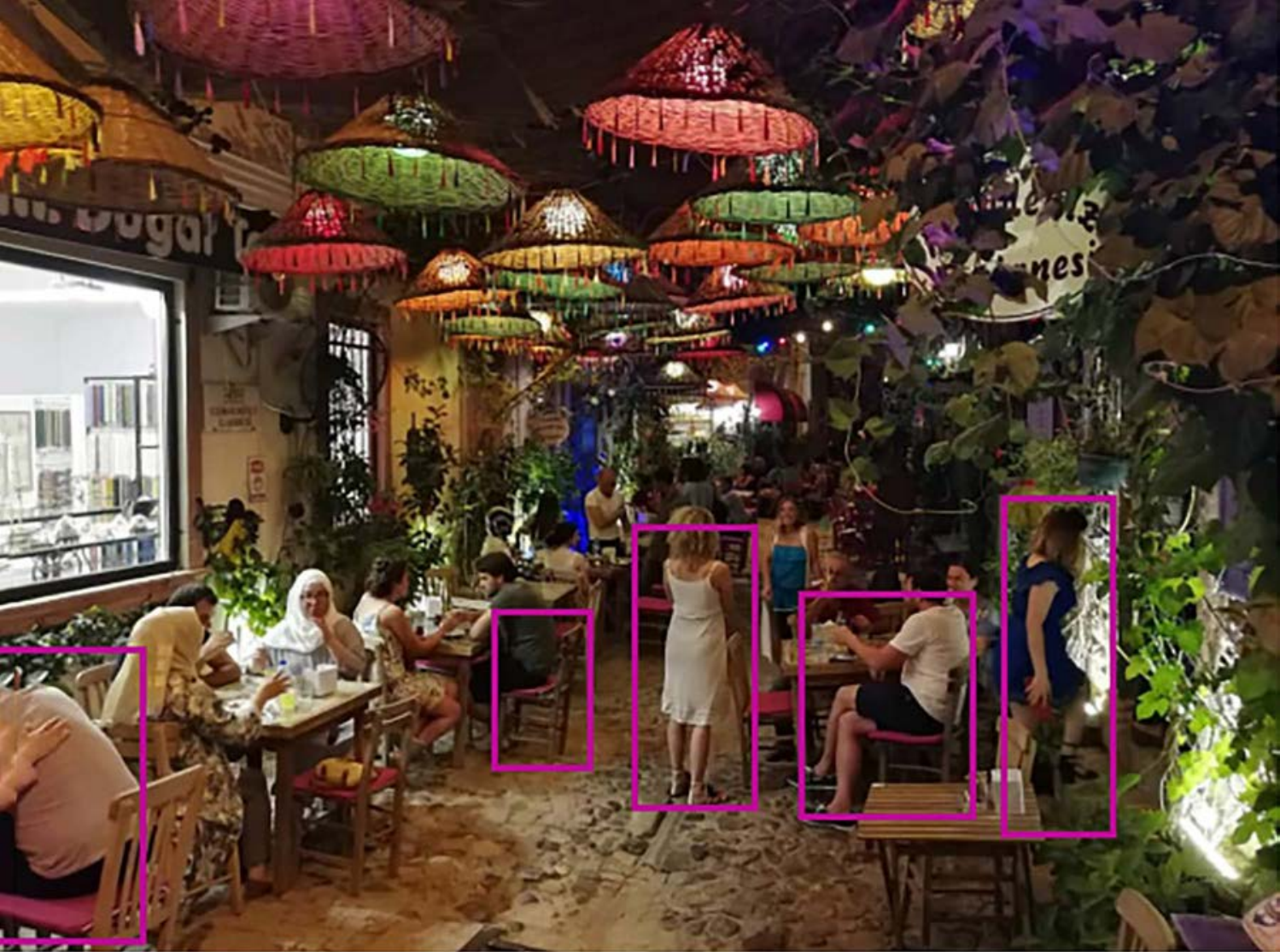}
        \label{fig:M3d}
    \end{subfigure}
    \hfill
    \begin{subfigure}[c]{0.4\textwidth}
        \includegraphics[width=\textwidth]{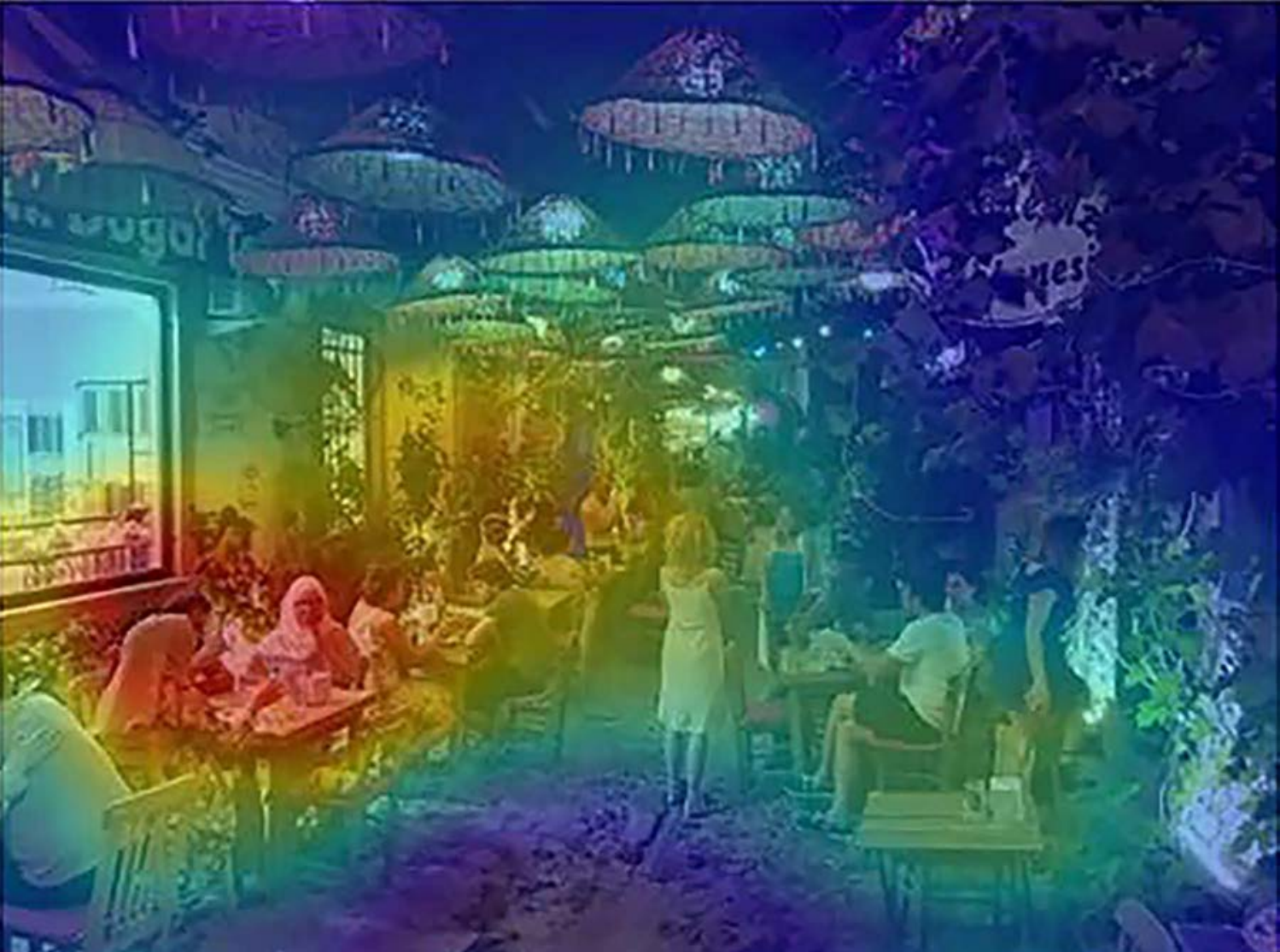}
        \label{fig:M3b}
    \end{subfigure}
    \hfill
    \begin{minipage}[c]{2.5in}
    \begin{trivlist}  
        \item {\normalsize top-1: indoor (0.978)\par}
        \item {\normalsize top-2: person (0.866)\par}
        \item {\normalsize top-3: people (0.619)\par}
    \end{trivlist}
    \subcaption{}
    \end{minipage}
    \hfill
    \begin{minipage}[c]{2.5in}
    \begin{trivlist}  
        \item {\normalsize top-1: restaurant (0.247)\par}
        \item {\normalsize top-2: cafeteria (0.236)\par}
        \item {\normalsize top-3: restaurant\textunderscore patio (0.172)\par}
    \end{trivlist}
    \subcaption{}
    \end{minipage}
\hfill
\caption{(a) Results provided by Microsoft Cognitive Services~\cite{microsoft,del2018introducing} with top provided tags, (b) the Places-CNN~\cite{zhou2018places} with the top provided results.
}
    \label{fig:M3}
\end{figure}

Another approach used to extract relevant clues and features from the images posted on SNS was utilizing the scene information as well as the presence of certain objects. We used two different pre-trained Deep Neural Networks for this purpose: Microsoft Azure Cognitive Services~\cite{microsoft,del2018introducing} to obtain information regarding the objects that appear in the pictures as well as tags related to the image, and Places-CNN~\cite{zhou2018places} for a description of the scene. Fig.~\ref{fig:M3} shows the results of applying these state-of-the-art methods.

Places-CNN is a AlexNet~\cite{krizhevsky2012imagenet} trained with Places365~\cite{zhou2018places}, which is the latest subset of Places2 Database. This network has been trained with more than 1.8 million images from 365 scene categories. In order to obtain information from the components of the scene, we also use the Microsoft Azure Cognitive Services~\cite{microsoft,del2018introducing} Computer Vision API, which provides a list of objects appearing in the image, relevant tags, and dominant colors.

To generate a representative feature vector for each subject to be used for classification, we used the information obtained by the CNNs to create a Bag of Words. Specifically, for the outputs of the Azure approach, we created a dictionary of tags, taking all the different tags detected in all the images of the database. In this case, the number of different tags is 734. Therefore, for each subject, we generated a histogram of occurrence of these tags in the photos that the subject has posted, obtaining a feature vector $x \in \mathcal{X}_{Azure} \subseteq \mathbb{R}^{734}$.

For the information gathered from the Places-CNN, we generated a similar histogram, obtaining a feature vector $x \in \mathcal{X}_{Places-CNN} \subseteq \mathbb{R}^{344}$.  However, since the results are exclusive from each other, we decided to weight the occurrences in the histogram according to the score obtained in AlexNet.

\subsection{Textual analysis}
\begin{figure}[htbp]
    \centering
    \begin{subfigure}[c]{0.245\textwidth}
        \includegraphics[width=\textwidth]{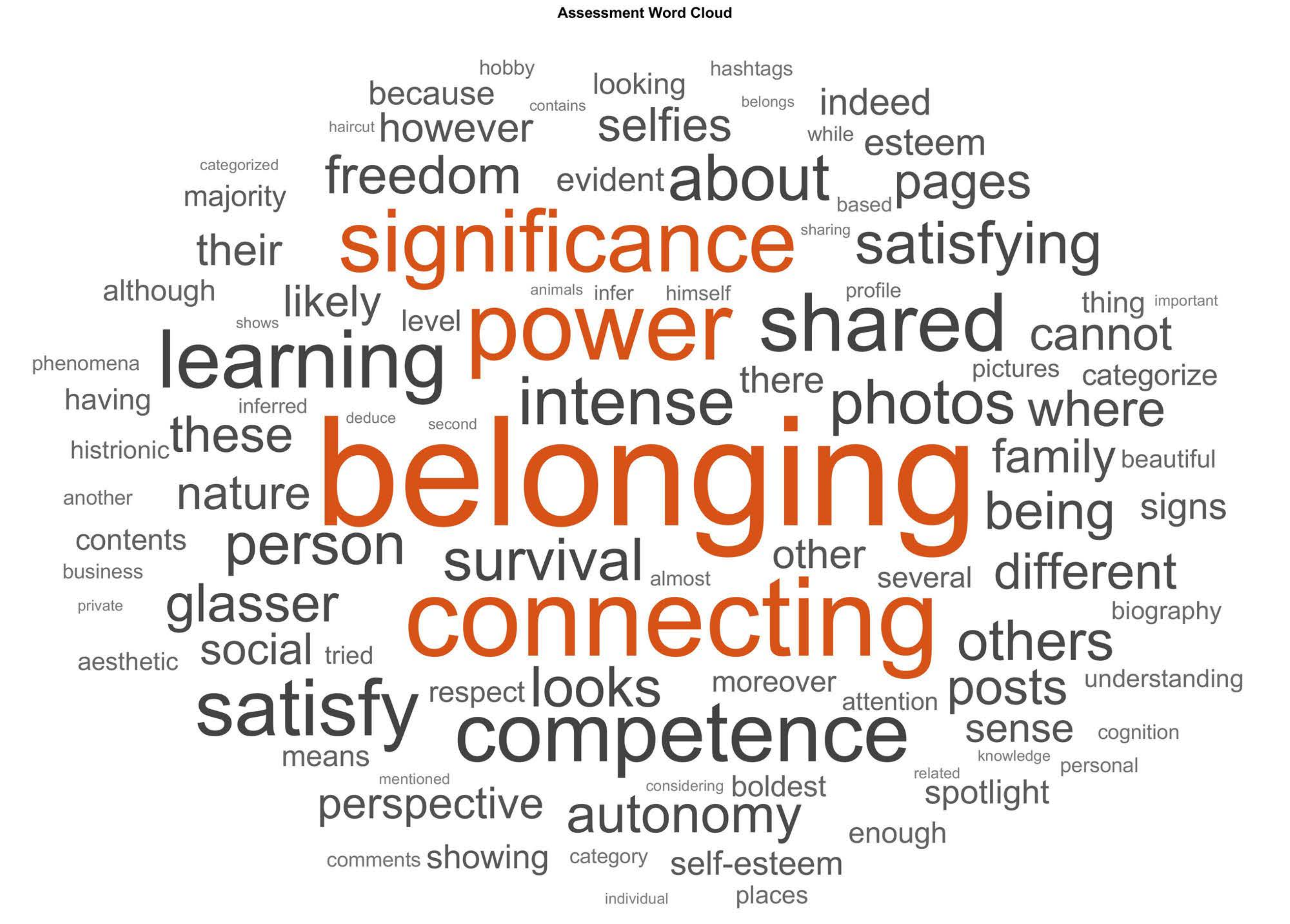}
        \caption{}
        \label{fig:M4a}
    \end{subfigure}
    \hfill
    \begin{subfigure}[c]{0.245\textwidth}
        \includegraphics[width=\textwidth]{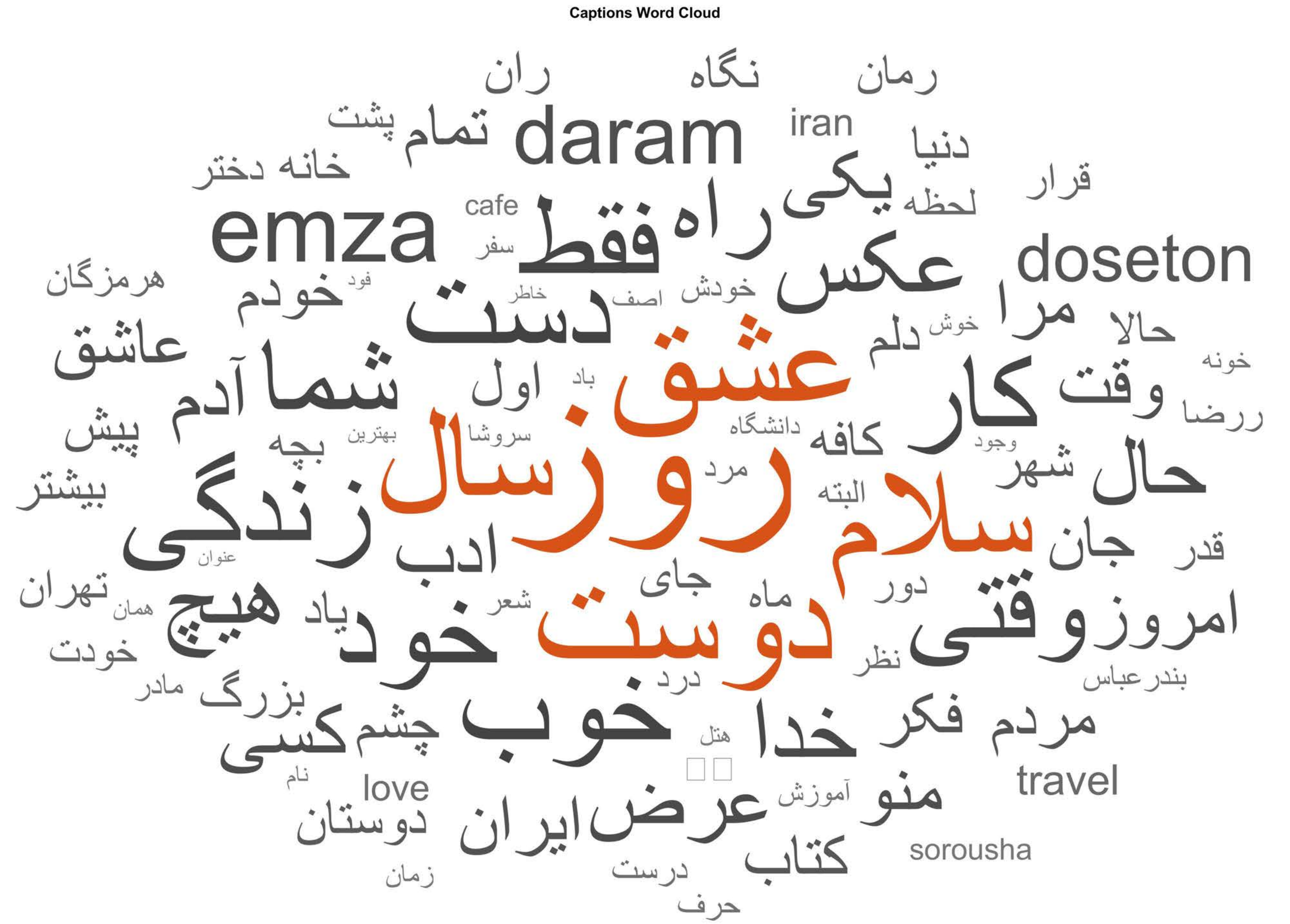}
        \caption{}
        \label{fig:M4b}
    \end{subfigure}
    \hfill
    \begin{subfigure}[c]{0.245\textwidth}
        \includegraphics[width=\textwidth]{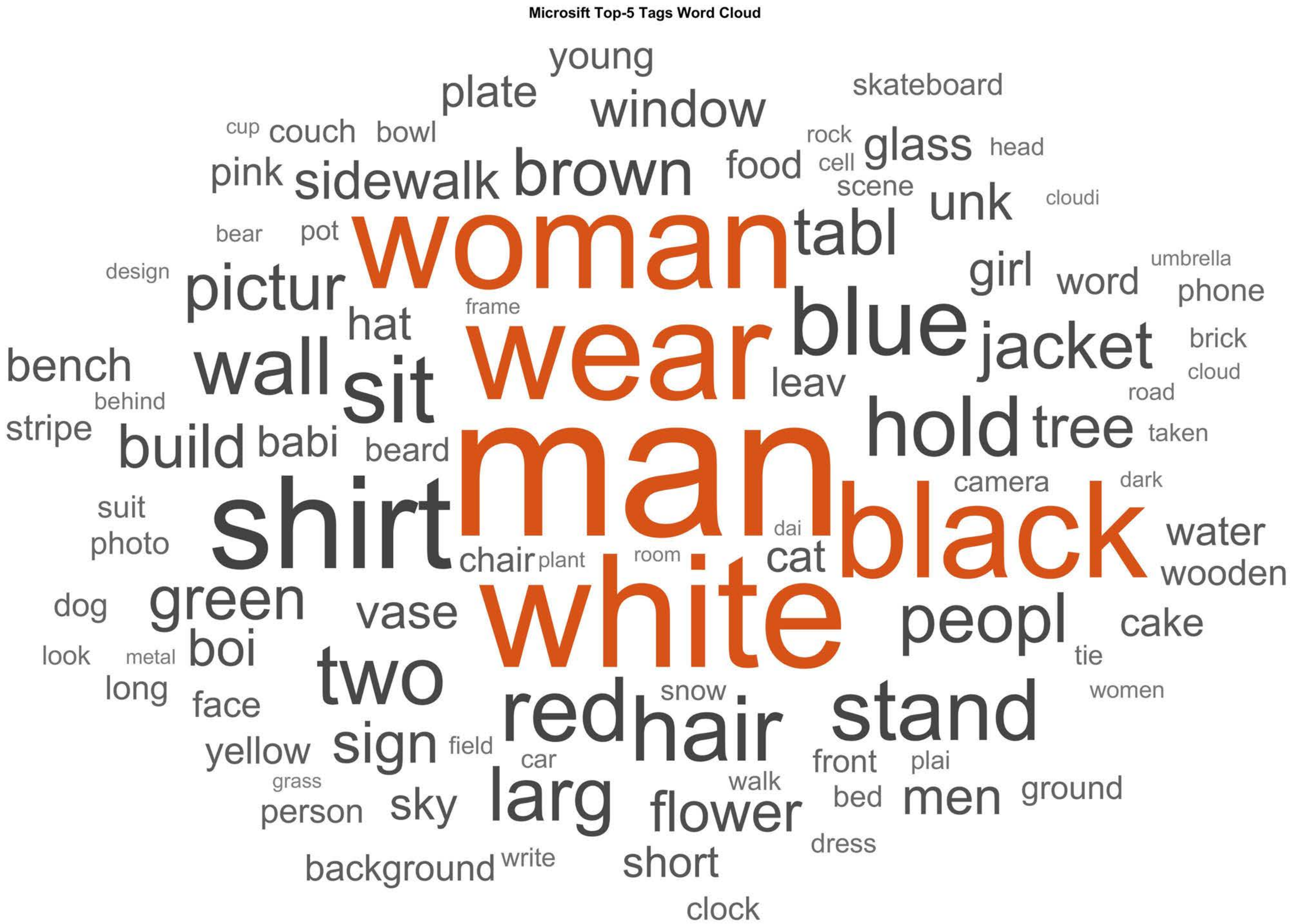}
        \caption{}
        \label{fig:M4c}
    \end{subfigure}
    \hfill
    \begin{subfigure}[c]{0.245\textwidth}
        \includegraphics[width=\textwidth]{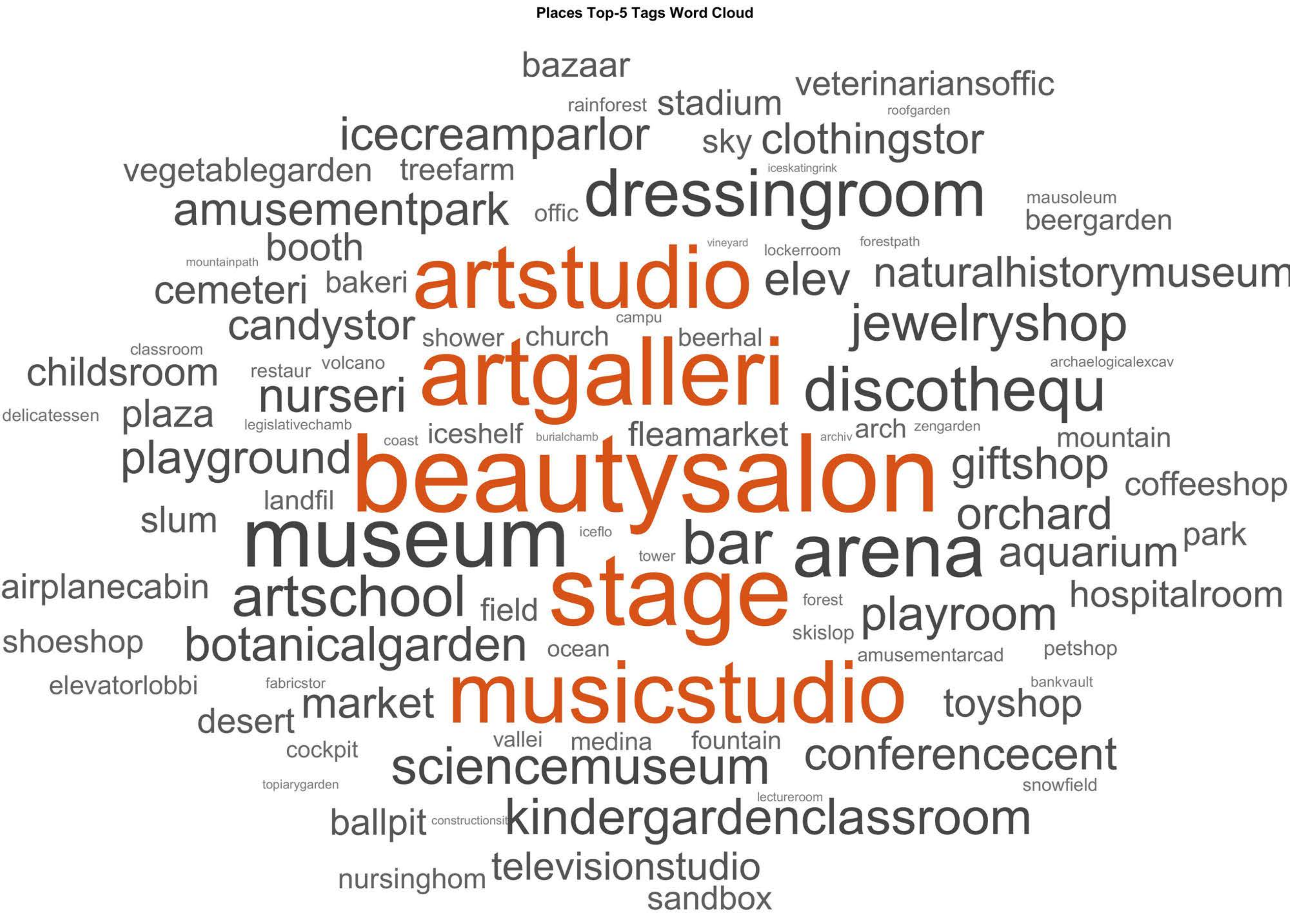}
        \caption{}
        \label{fig:M4d}
    \end{subfigure}
    \hfill
    \begin{subfigure}[c]{0.245\textwidth}
        \includegraphics[width=\textwidth]{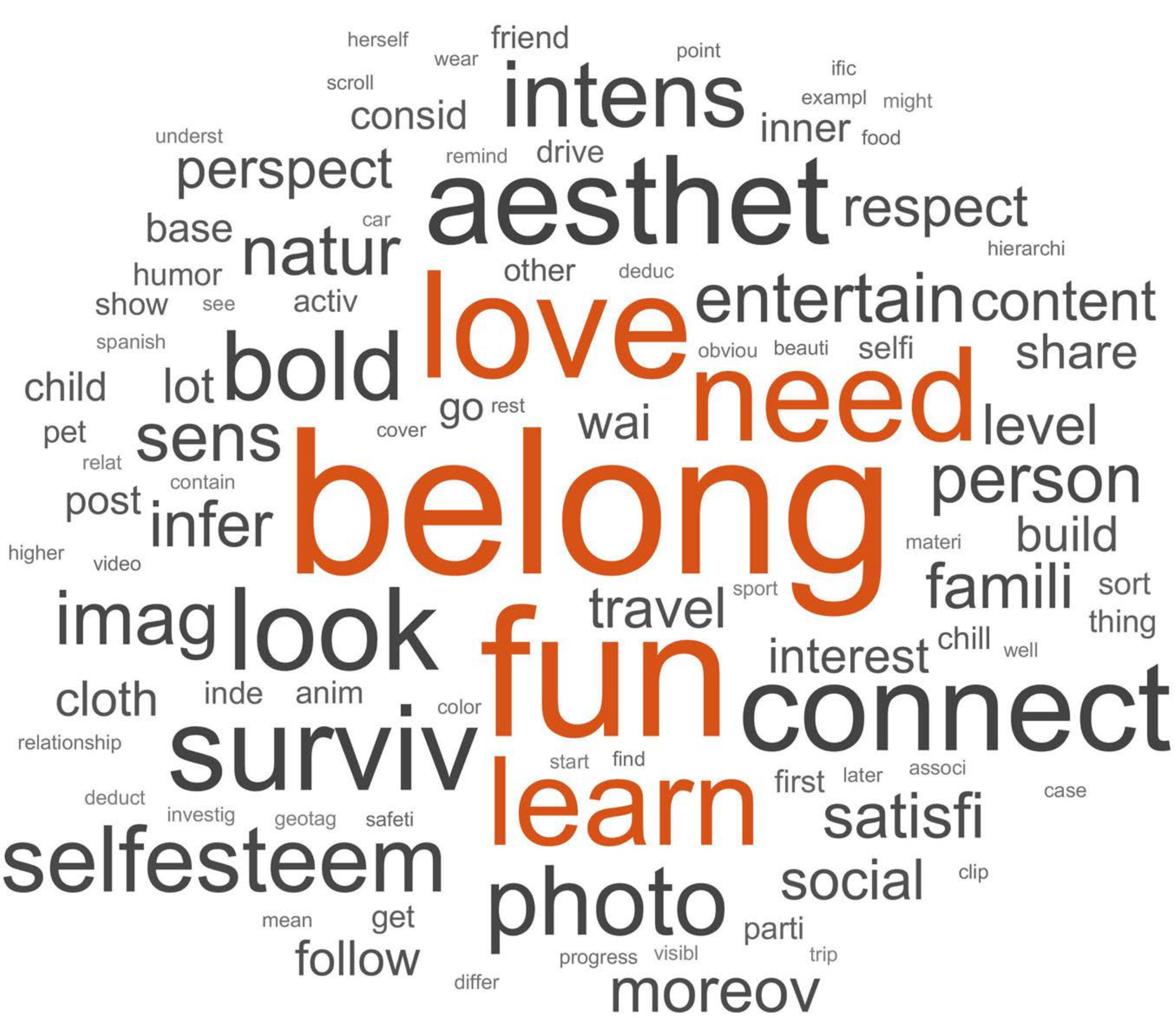}
        \caption{}
        \label{fig:M4e}
    \end{subfigure}
    \hfill
    \begin{subfigure}[c]{0.245\textwidth}
        \includegraphics[width=\textwidth]{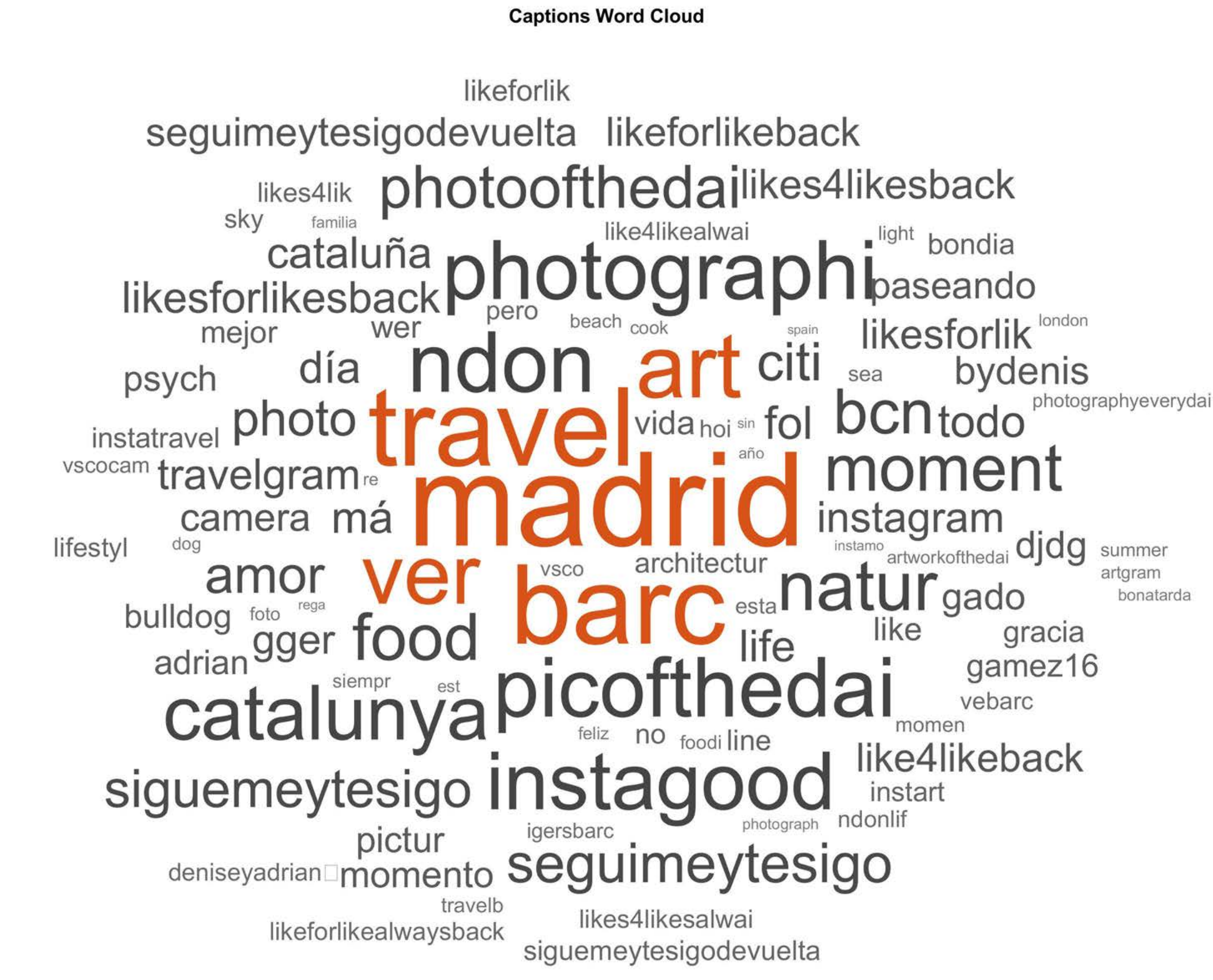}
        \caption{}
        \label{fig:M4f}
    \end{subfigure}
    \hfill
    \begin{subfigure}[c]{0.245\textwidth}
        \includegraphics[width=\textwidth]{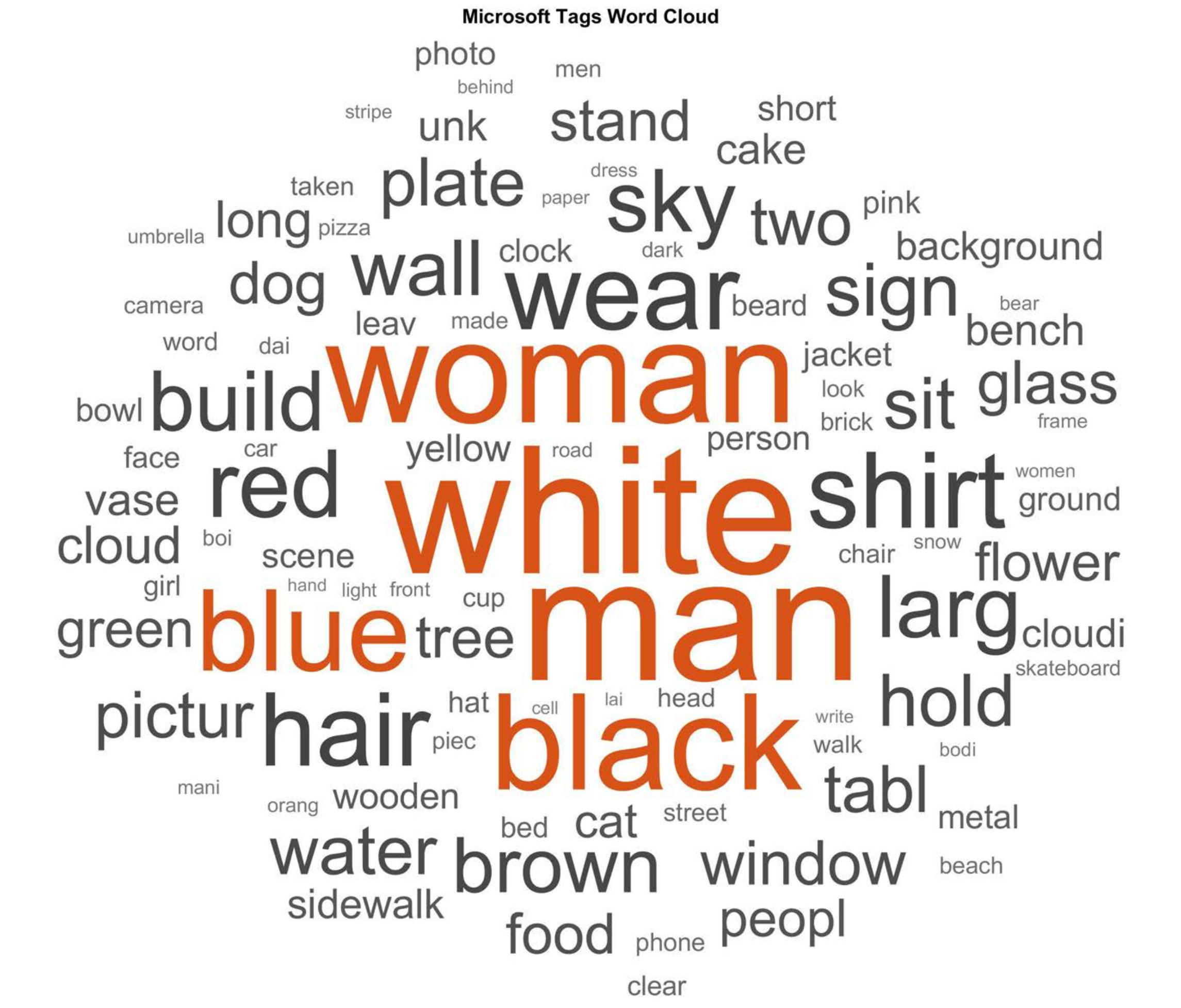}
        \caption{}
        \label{fig:M4g}
    \end{subfigure}
    \hfill
    \begin{subfigure}[c]{0.245\textwidth}
        \includegraphics[width=\textwidth]{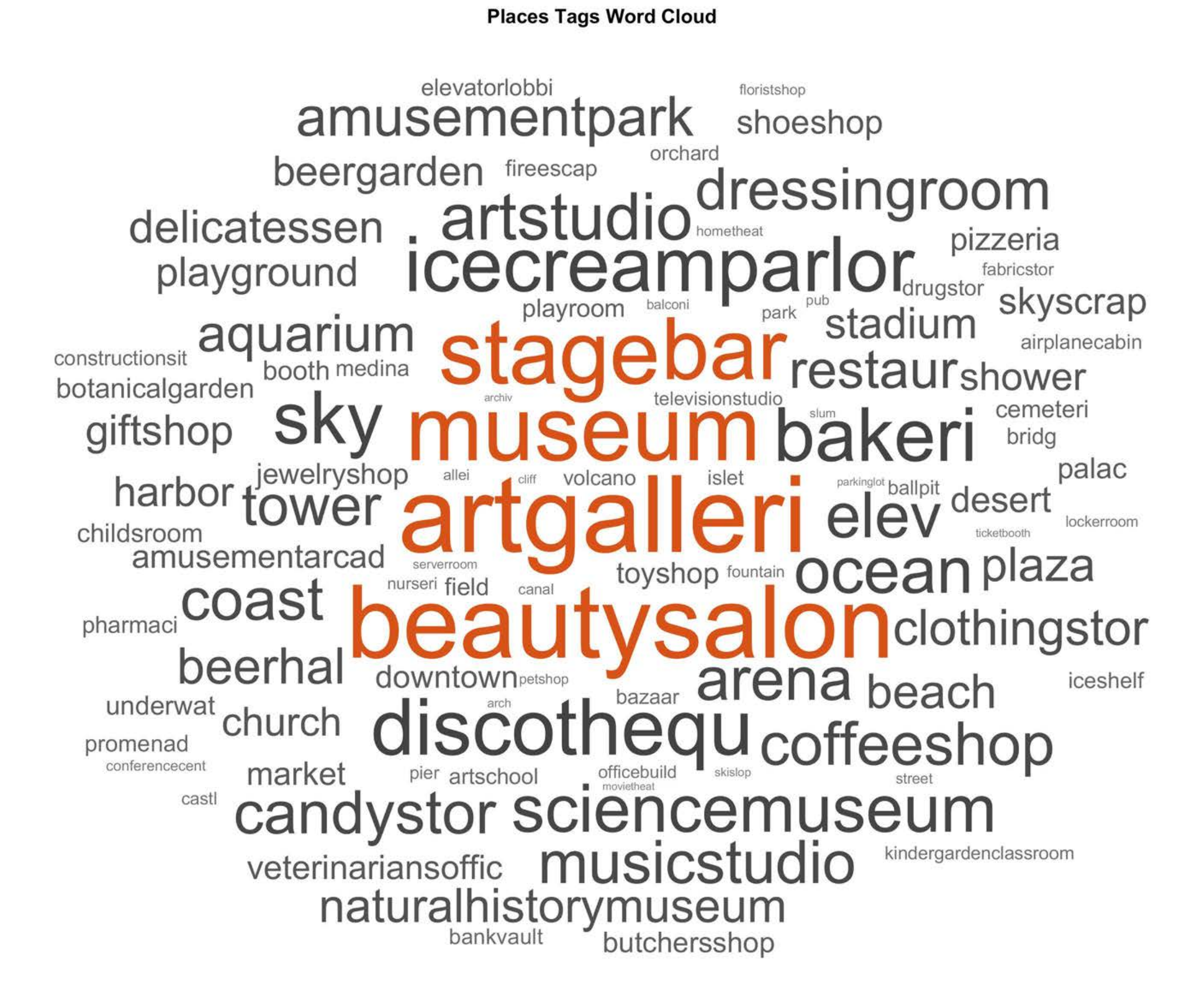}
        \caption{}
        \label{fig:M4h}
    \end{subfigure}
    \caption{The word cloud of (a,e) psychologist deductions, (b,f) all captions provided by Instagram users, (c,g) tags provided by Microsoft Cognitive Services, (d,h) tags provided by Place network applied to images of Instagram profile. [Top row] Iran profiles, [Bottom row] Spain profiles.}
    \label{fig:M4}
\end{figure}

To investigate in textual information provided by users and understand its importance, we first created a word cloud model for each textual information which records the number of times that words appear in each document in a collection. Figures~(\ref{fig:M4a}-\ref{fig:M4e}) shows the top words used by the expert Psychologist describing all profiles, Figures~(\ref{fig:M4b}-\ref{fig:M4f}) show the word clouds from textual captions, Figures~(\ref{fig:M4c}, \ref{fig:M4g}, \ref{fig:M4d}, and~\ref{fig:M4f}) show the word clouds obtained from tags provided by visual object detectors. The first row is associated with Iranian profiles and the second row with Spanish profiles. Psychologist descriptions literally fit better with textual captions provided by users. For instance, one can see that the most intense word in Fig.~\ref{fig:M4a} is `belonging' where the top-5 frequent words are \{`belonging', `page', `love', `connect', `satisfy'\}. 

In Fig.~\ref{fig:M4b}, we have the cloud of captions which the boldest word is \raisebox{-.35\height}{\includegraphics[width=0.9cm]{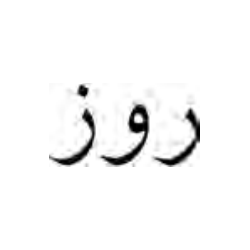}}~\footnote{The meaning in English is `day'.} and the top-5 words translated into English are \{`day', `love', `eye', `friend', `window'\}. Comparing these two sets, we can associate friend and love to `belonging'. However, the cases for Fig.~\ref{fig:M4c} and~\ref{fig:M4d} are almost different. The top-5 words in these figures are \{`beautysalon', `stage', `artgalleries', `artstudio', `musicstudio'\}, \{`man', `white', `wear', `woman', `black'\}, respectively, which are not literally related to `Belonging/connecting/love,'. 
 
In order to use the textual information provided by the users in their SNS posts, we followed a similar approach to the visual data. We generated a feature vector for each user to be classified in a further step. The feature vector is the result of training a Word2vec~\cite{word2vec} network, which learns an embedding representation of a dataset of words. Therefore, all the tokenized words extracted from all the users' posts were used to train the network. Note that given the differences in Persian and Spanish languages, we trained two different networks. Once trained, the words of each user were passed through the network to obtain their embedded representation. Finally, all the embedded vectors were averaged in order to obtain an unique feature vector per user, $x \in \mathcal{X}_{text} \subseteq \mathbb{R}^{128}$.

\subsection{Multi-label classification}
The natural way to tackle this problem is to follow a multi-label classification rule. For example, a psychologist can perceive from a user profile that the individual tries to satisfy Power, Freedom, and Fun needs simultaneously. Therefore, we decided to use Multi-Label Learning with GLObal and loCAL Label Correlation (GLOCAL) method~\cite{zhu2018multi}.

Let $C=\{c_{1}, \cdots, c_{l}\}$ be the set of $l$ class labels. The $d$-dimensional feature vector of an instance is denoted by $x \in \mathcal{X} \subseteq \mathbb{R}^{d}$, and the ground-truth label vector is denoted by $ \tilde{y} \in \mathcal{Y} \subseteq \{-1,1\}^{l}$, where $[\tilde{y}]_{j}=1$ if $x$ is with class label $c_{j}$, and -1 otherwise. GLOCAL classifier provides outputs by solving the following optimization (Eq.~\ref{eq:M1}). The outputs are encouraged to be similar on highly positively correlated labels, and dissimilar on highly negatively correlated labels.

\begin{equation} \label{eq:M1}
    \begin{aligned}
        \min_{U,V,W,\mathcal{Z}} & \parallel J \circ (Y-UV) \parallel^{2}_{F} + \lambda \parallel V-W^{T}X \parallel^{2}_{F} +\\
        & \sum_{m=1}^{g} \left ( \frac{\lambda_{3}n_{m}}{n}\mathrm{tr}(F^{T}_{0}Z_{m}Z^{T}_{m}F_{0}) + \lambda_{4}\mathrm{tr}(F^{T}_{m}Z_{m}Z^{T}_{m}F_{m}) \right ) + \lambda_{2}\mathcal{R}(U,V,W) &&\\
        \text{subject to} &&\\
        & \text{diag}(Z_{m}Z^{T}_{m}) = 1, m=1, 2, \cdots, g. &&
    \end{aligned}
\end{equation}
where $\lambda, \lambda_{2}, \lambda_{3}, \lambda_{4}$ are trade-off parameters, $\mathcal{R}(U,V,W)$ is a regularizer, $F$ is the vector containing predictions on all $n$ instances, $J=[J_{ij}]$ is the indicator matrix, $V$ represents the latent labels and $U$ reflects how the original labels are correlated to the latent labels. $Y$ is the observed labels and $W \in \mathbb{R}^{d \times k}$ is used to map instances to the latent labels. The $W$ can be obtained by minimizing the square $\parallel V-W^{T}X \parallel^{2}_{F}$, where $X=[x_{1}, \cdots, x_{n}] \in \mathbb{R}^{d \times n}$ is the matrix containing all the instances. $L$ is a learning Laplacian matrix which preserves the label correlation and can be written as a learning $\mathcal{Z} \equiv \{z_{1}, \cdots, Z_{g}\}$ for $m \in \{1, \cdots, g\}$. $\text{tr}(\bullet)$ is the trace of $\bullet$, $\parallel \bullet \parallel_{F}$ is its Frobenius norm, $\text{diag}(\bullet)$ returns a vector containing the diagonal elements of $\bullet$. For two matrices of the same size, A and B, $A \circ B$ denotes the Hadamard (element-wise) product.

\section{Experiments} \label{experiment}
We used cross-validation (leave one subject out) approach to evaluate the GLOCAL model due to the limited number of the data samples. Therefore, we selected $N-1$ of the instances for training, and the remaining for testing. This validation approach repeats $N$ times for two subsets and results are averaged over $N$ independent repetitions to reduce statistical variability. The method received as input the four feature spaces used in this paper (BoVW, Azure, Places-CNN and text). Due to the fact that the expert psychologist uses both information from images and captions when carrying out the psychological assessment, we also included a Fusion feature space where all the generated feature spaces were concatenated and normalized.

Table~\ref{tbl:M1} reports the evaluation criteria which are Average precision (Ap), Area under the ROC curve (Auc),  Hamming loss (Hl), and Jaccard similarity score (Jsc).

\begin{itemize}
    \item Average precision (Ap) summarizes a precision-recall curve as the weighted mean of precision values achieved at each threshold, with the increase in recall from the previous threshold used as the weight: $\text{Ap} = \sum_{i} (R_i - R_{i-1}) P_i $ where $P_i$ and $R_i$ are the precision and recall at the i-th threshold.
    \item Area under the ROC curve (Auc) defines the area under the plot of the fraction of true positives out of the positives (TPR = true positive rate) vs. the fraction of false positives out of the negatives (FPR = false positive rate), at various threshold settings.
    \item Hamming loss (Hl) is the fraction of labels that are incorrectly predicted. Hamming loss defines as $\text{Hl}=\frac{1}{LN}\sum_{l=1}^{L}\sum_{i=1}^{N} \tilde{y}_{i,l} \oplus f(x_{i,l})$, where $\oplus$ is exclusive-or, $L=\text{card}(C)$, $N$ is the number of instances. $\tilde{y}_{i,l}$ and $f(x_{i,l})$ stand for Boolean that, in turn, the $i$-{th} data and prediction contains the $l$-{th} label.
    \item Jaccard similarity score (Jsc) is the size of the intersection of the predicted labels $f(x)$ and the true labels $\tilde{y}$ divided by the size of the union of the predicted and true labels. Jsc is given by $\text{Jsc}(f(x),\tilde{y})=\frac{f(x) \cap \tilde{y}}{f(x) \cup \tilde{y}}$
\end{itemize}

Moreover we calculated Ranking loss and Coverage as in \cite{zhou2012multi}. Let $p$ be the number of test instances, $C_{i}^{+}, C_{i}^{-}$ be the sets of positive and negative labels associated with the $i$-{th} instance; and $Z_{j}^{+}, Z_{j}^{-}$ be the sets of positive and negative instances belonging to the $j$-{th} label. Given input $x$, let $\text(rank)_{f}(x,y)$ be the rank of label $y$ in the predicted label ranking (sorted in descending order).

\begin{itemize}
    \item Ranking loss (Rkl): This is the fraction that a negative label is ranked higher than a positive label. For instance $i$, define $\mathcal{Q}_{i} = \{(j^{'},j^{''})|f_{ j^{'}}(x_{i}) \leq f_{ j^{''}}(x_{i}), (j^{'},j^{''}) \in C_{i}^{+} \times C_{i}^{-}\}$. Then, $\text{Rkl} = \frac{1}{p} \sum_{i=1}^{p}\frac{|\mathcal{Q}_{i}|}{ |C_{i}^{+}| |C_{i}^{-}|}$.
    \item Coverage (Cvg): This counts how many steps are needed to move down the predicted label ranking so as to cover all the positive labels of the instances. $\text{Cvg} = \frac{1}{p} \sum_{i=1}^{p} \max\{\text{rank}_{f}(x_{i},j) | j \in C_{i}^{+}\}-1$.
\end{itemize}

For $\text{Auc}$, $\text{Ap}$ and $\text{Jsc}$, the higher are the better; whereas for $\text{Rkl}$, $\text{Cvg}$ and $\text{Hl}$, the lower are the better.

\begin{table}[htbp]
\centering
\caption{Results of applying GLOCAL to public profiles of Iran (42 users) and Spain (46 users) Instagram considering ranking loss (Rkl), average area under curve (Auc), coverage (Cvg), average precision (Ap), Hamming loss (Hl), and Jaccard similarity score (Jsc).`\#dim' is the dimension of feature vector. We show the mean measurement with $95\%$ confidence intervals.}
\label{tbl:M1}
\resizebox{\textwidth}{!}{%
\begin{tabular}{lll|llllll}
\hline
\textbf{} & \textbf{Feature Space} & \textbf{\#dim} & \textbf{Rkl} & \textbf{Auc} & \textbf{Cvg} & \textbf{Ap} & \textbf{Hl} & \textbf{Jsc} \\ \hline
\multirow{5}{*}{\textbf{Iran}} & BoVW & 256 & $\mathbf{0.079 \pm 0.01}$ & $0.778 \pm 0.08$ & $\mathbf{1.549 \pm 0.05}$ & $0.814 \pm 0.07$ & $0.285 \pm 0.07$ & $0.714 \pm 0.08$ \\
 & Places-CNN & 344 & $0.086 \pm 0.02$ & $0.671 \pm 0.12$ & $1.574 \pm 0.12$ & $0.693 \pm 0.12$ & $0.406 \pm 0.09$ & $0.593 \pm 0.09$ \\
 & Azure & 734 & $\mathbf{0.079 \pm 0.01}$ & $0.760 \pm 0.09$ & $1.552 \pm 0.08$ & $0.756 \pm 0.11$ & $0.351 \pm 0.07$ & $0.648 \pm 0.07$ \\
 & Word2Vec & 128 & $0.145 \pm 0.04$ & $\mathbf{0.805 \pm 0.08}$ & $1.808 \pm 0.31$ & $ \mathbf{0.835 \pm 0.07}$ & $\mathbf{0.250 \pm 0.07}$ & $\mathbf{0.750 \pm 0.07}$ \\
 & Fusion & 1462 & $0.142 \pm 0.06 $ & $0.803 \pm 0.09$ & $ 1.768 \pm 0.33 $ & $0.785 \pm 0.10$ & $0.339 \pm 0.07$ & $0.660 \pm 0.07$ \\ \hline
\multirow{5}{*}{\textbf{Spain}} & BoVW & 256 & $0.060 \pm 0.01$ & $0.927 \pm 0.05$ & $\mathbf{1.426 \pm 0.06}$ & $\mathbf{0.928 \pm 0.05}$ & $0.156 \pm 0.05$ & $0.843 \pm 0.05$ \\
 & Places-CNN & 344 & $0.063 \pm 0.01$ & $0.789 \pm 0.11$ & $1.504 \pm 0.06$ & $0.805 \pm 0.09$ & $0.281 \pm 0.07$ & $0.718 \pm 0.07$ \\
 & Azure & 734 & $\mathbf{0.051 \pm 0.01}$ & $0.875 \pm 0.07$ & $1.442 \pm 0.05$ & $0.913 \pm 0.05$ & $0.250 \pm 0.08$ & $0.750 \pm 0.08$ \\
 & Word2Vec & 128 & $0.064 \pm 0.03$ & $0.927 \pm 0.04$ & $1.442 \pm 0.20$ & $0.889 \pm 0.07$ & $0.200 \pm 0.05$ & $0.800 \pm 0.05$ \\
 & Fusion & 1462 & $0.064 \pm 0.04 $ & $\mathbf{0.933 \pm 0.05}$ & $ 1.491 \pm 0.250 $ & $0.917 \pm 0.07$ & $\mathbf{0.143 \pm 0.06}$ & $\mathbf{0.856 \pm 0.06}$ \\ \hline
\end{tabular}%
}
\end{table}

As we can see in Table~\ref{tbl:M1}, classification using features generated with Word2vec from the textual information of captions outperforms the rest of the methods using most of the metrics for the case of users from Iran. For Spanish users the fusion of features from images and text works the best. These findings show the relevance of the textual information in the classification. Therefore, one could say that the information shared by the users in the captions of the photos are important cues for perceiving their pursuit of needs.

Among the methods using information from images, SURF-based BoVW is the one with the best results. Since these features are extracted directly from the images, they represent better their information. On the other hand, features obtained with Places-CNN or Azure are generated from the occurrence of objects and tags at a higher level, missing features present on the images themselves. We conjecture that more sophisticated models of scene understanding that could take into account high level relationships between objects could increase the performance of these descriptors. Similarly, a more robust captioning about the scene contents (specifically regarding to the nature of the interactions among the people present in the scene) would improve the predicted perception of needs.

\section{Conclusion} \label{conclusion}
We introduced the VICSOM database, a multimodal database of 86 public Instagram accounts, containing 30,080 images, assessed by an expert psychologist with a focus on human needs. We considered gender and age diversities in harvesting profiles where the subjects belong to the age interval of 15-50 with a gender ratio (male/female) of 1.68. To perceive needs, the expert psychologist took the Glasser's choice theory into account in which it is stated that human behaviors are driven by five genetically driven needs including survival, love and belonging, freedom, fun, and power. The VICSOM DB has also been made publicly available to the research community, representing a benchmark for efforts in automatic categorizing human needs over Instagram as a trending social network site (SNS). 

We provided exhaustive baseline experiments to assess textual/visual features in advancing the field of multimodal SNS analysis which would help in screening mental health. In the line of experiments, a multi-label classifier was trained and evaluated by three different feature representation methods which are (1) a bag of visual words formed by SURF descriptor, (2) a histogram of visual tags provided by two different Convolutional Neural Networks, and (3) textual descriptors by creating vectors that are distributed numerical representations of word features using Word2vec. We also explored a multimodal fusion of both visual and textual cues for the multi-label classification. We believe this data corpus will be helpful to the community, both in the psychological field in helping test hypothesis and in the computer science field to advance the state of automatic SNS analysis. 

We observed that the subjects' needs had experienced an evolution during the time. Therefore, one open research line is to investigate this evolution rather than considering static information at the moment of analysis. Furthermore, users of a specific SNS usually have profiles in other social networking sites which can be used for the screening together. Since the Instagram introduced `live' and `story' features, the taste of users is also shifting into sharing this kind of posts. In this way, further study can be performed using the visual and textual information from this data. In addition, further developments could include the use of other modalities, such as data from wearable devices (accelerometers, global localization) or regular communications (email, blogs, ...).

We noticed that scene description algorithms lack in finding the emotional state of a scene in which humans have social interactions. Indeed, not only objects, scenes, and sentiments but also relationships among scene components could be considered. For instance, if users shared images of themselves in a family gathering with intimate partners, it is likely to perceive that they look for a way to satisfy the need of belonging. However, if the image shows the interaction of users with religious groups, the inference about their need could be different.

There are other categorizations of human needs that could be explored in the future. For instance, one can consider the Transactional theory proposed by Eric Berne et al. ~\cite{berne1996transactional} in which the principal characters are child, parent, and mature. We also plan as a future works to explore more robust fusion rules for multimodal exploration and the use of alternative multilabel classifiers.

\section*{Acknowledgements}
This research was supported by TIN2015-66951-C2-2-R, RTI2018-095232-B-C22 grant from the Spanish Ministry of Science, Innovation and Universities (FEDER funds), and NVIDIA Hardware grant program.

\section*{Additional Information}
Implementations are available at https://github.com/dehshibi/VICSOM

\bibliographystyle{spbasic}
\bibliography{references}

\end{document}